\documentclass[10pt]{article}

\usepackage{graphics}
\usepackage{graphicx}

\usepackage[a4paper, left=35mm,right=35mm,top=34mm,bottom=34mm]{geometry}
\usepackage[utf8]{inputenc}
\usepackage[T1]{fontenc}
\usepackage[english]{babel}

\usepackage{enumerate}
\usepackage{graphicx}
\usepackage{hyperref}
\hypersetup{
    colorlinks=true,
    linkcolor=blue,
    filecolor=magenta,      
    urlcolor=cyan,
}
\usepackage{listings}
\usepackage{color}

\definecolor{dkgreen}{rgb}{0,0.6,0}
\definecolor{gray}{rgb}{0.5,0.5,0.5}
\definecolor{mauve}{rgb}{0.58,0,0.82}

\usepackage{mathtools,amsthm,amssymb,amsfonts}
\usepackage{algorithm}
\usepackage{algorithmic}
\makeatother
\theoremstyle{plain}

\theoremstyle{definition}

\theoremstyle{remark}

\usepackage{caption} 
\usepackage{fancyhdr}
\title{A Hybrid GNN approach for predicting node data for 3D meshes.\thanks{GNN: graph neural network}}

\author{Shwetha Salimath \thanks{Universit\'e Paris-Saclay, CentraleSupelec, France}\and
Francesca Bugiotti \thanks{Universit\'e Paris-Saclay, CentraleSupelec, LISN, France}\and
Fr\'ed\'eric Magoul\`es \thanks{Universit\'e Paris-Saclay, Centralesuplec, MICS, France}}

\date{}

\begin{document}
\maketitle 
\thispagestyle{fancy}

\begin{abstract}
Metal forging is used to manufacture dies. We require the best set of input parameters for the process to be efficient. Currently, we predict the best parameters using the finite element method by generating simulations for the different initial conditions, which is a time-consuming process.
In this paper, introduce a hybrid approach that helps in processing and generating new data simulations using a surrogate graph neural network model based on graph convolutions, having a cheaper time cost. 
We also introduce a hybrid approach that helps in processing and generating new data simulations using the model. Given a dataset representing meshes, our focus is on the conversion of the available information into a graph or point cloud structure. This new representation enables deep learning. The predicted result is similar, with a low error when compared to that produced using the finite element method. The new models have outperformed existing PointNet and simple graph neural network models when applied to produce the simulations.
\end{abstract}


\section{Introduction}

Metal forging is the process used to shape metals using compressive forces, and multiple parameters influence this process. Hot forging seals minor cracks and redistributes impurities leading it to be the most used in the industry.
But this process has a high cost associated with the manufacturing of the forging die. This is due to the need of setting the production environment by tuning the important initial parameters by multiple iterations. In the beginning, tuning was done by producing samples resulting in lots of energy, time, and material wastage. This led to the use of simulation software to provide the best set of initial parameters for product manufacturing.

The Finite Element Method (FEM) has been used as a significant part of designing feasible metal forging processes~\cite{behrens2008finite}. The FEM calculates and gives us the simulation depending on the set of feasible conditions to select the best set of input parameters. The objective of FEM~\cite{harari2004numerical} is to solve partial differential equations resulting in a system of algebraic equations. Large meshes require a lot of computational time and resources for optimizing and running the process, attaching a very high time cost to the product.

 Artificial neural networks~\cite{hassoun1995fundamentals} (ANN) are machine learning models used mostly for solving problems on conventional regression and statistical models~\cite{abiodun2018state}. The graph neural network (GNN) model~\cite{4700287} allows the processing of the data represented as a graph. They have two main purposes, (i) graph-focused, or (ii) node-focused. 

In this paper, we propose a hybrid approach that uses FEM and a deep learning model together to create quicker simulations for finding the best set of input conditions. In this hybrid approach, we introduce a GNN model which is trained on a dataset of meshes. These meshes are generated by using FEM simulations for a subset of initial conditions. The trained model is then used to predict the simulations for the rest. Once trained, the model can generate one simulation in 300 milliseconds, while the FEM would take about 45 minutes. The proposed model would thus be 99.9\% faster than the FEM software. The simulations generated from the trained model, though not completely accurate, are good. We achieve an average mean absolute error of 10 Newton/meter at a mesh point, for which the actual wear ranges from 0-2000 N/m. 

The paper is organized as follows.  In Section~\ref{sec:relatedwork} we discuss the literature review. This is followed by Section~\ref{sec:methodology}, which explains the whole process in detail. In Section~\ref{sec:results} we compare our models with baseline models and end with a conclusion and future work in Section~\ref{sec:Conclusion}. 

\section{Related Work}
\label{sec:relatedwork}

There has been a rapid development in using 3D data for deep learning as it has numerous applications in different domains like robotics, autonomous driving, medical, and analyzing 3D objects in manufacturing industries. We can represent the 3D data as a point cloud, meshes, depth images, or grids. 

Point clouds have been the most popular form of 3D representation. PointNets~\cite{qi2017pointnet} are ANN used as a baseline for classification and segmentation tasks of 3D objects. The PointNet model upscales and then downscales the point cloud features using 1D convolution layers with activation function and max pooling.

Solving a FEM simulation is a difficult task. Attempts have been made to solve the partial differential equations using deep learning models known for their powerful function-fitting capabilities~\cite{guo2020solving}. These are in the field of biomechanics to simulate phenomena in anatomical structures~\cite{phellan2021real}.

For applications related to automated analysis of the generated meshes, PointNet, and GNN like MeshNet and graph convolution networks are slowly being introduced, once trained, are efficient and time-saving. MeshNet~\cite{feng2019meshnet} is used to learn features from the mesh data, which helps to solve the irregularity problems in representing the shape.

 Graph convolution layers~\cite{kipf2016semi} use neighbor degree and node features and scale them linearly in terms of the number of graph edges to learn hidden layer representation. The graph features can be extracted without the need to perform extra transformations. The Edge convolution layer~\cite{wang2019dynamic} uses the k-nearest neighbor of patch centers for constructing sequential embedding by extracting global features and pairwise operations for local neighborhood information. The SAGE convolution layer~\cite{hamilton2017inductive} uses sampling and aggregation of features from the neighborhood. Thus, with each iteration, more information is gained due to the aggregation, which could be mean, pooling, or graph convolution function. All of these methods were used for the classification of graphs or segmentation of graph nodes and not in generating simulations for meshes or graphs.

\section{Methodology}
\label{sec:methodology}
The main objective is to create simulations as produced by the FEM for a new set of initial conditions in metal forging. The output parameter is wear at each node of the mesh, which tells us about the damage caused in the forging die during the process.

\begin{figure*}[h]
    \centering
    \includegraphics[width=12cm, height = 4.5cm]{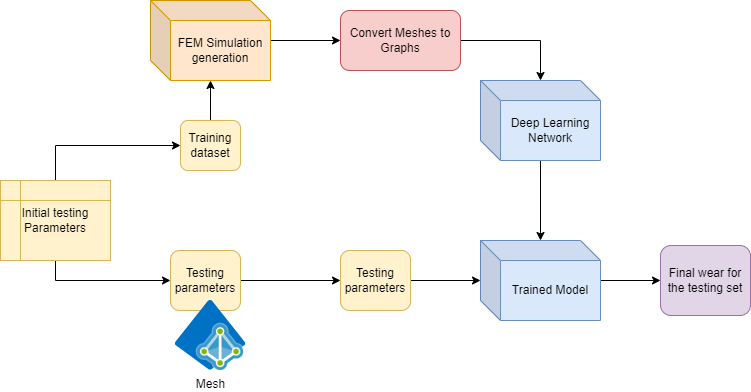}
    \caption{Block diagram of the process}
    \label{Fig 1}
\end{figure*}

We start with the set of initial conditions as parameters.
A very small subset is used to generate FEM simulations which will be used for training the GNN model. We need to extract mesh node data to create a graph or point cloud~\cite{rusu20113d}, used in deep learning models. Once the model is trained, we pass on the rest of the set to get the simulations . The process is represented in Fig~\ref{Fig 1}.

\subsection{Dataset}

The die designs of the Yoke metal forging process are provided as data by Transvalor. Mesh is composed of cells and points. We work with unstructured meshes, having sparse or arbitrary cell numbering within the mesh. We use Transvalor packages to produce an  unstructured mesh from the FEM simulations. These meshes are then analysed and converted into graphs or point clouds by extracting information using pyvista.

Each cell of the mesh has information such as temperature, pressure, displacement stress, etc. stored in them.  We only require the "wear" which is our output feature to be predicted. Cells can be of two types, 2D or 3D. The meshes in our case are made of 3D tetrahedron cells. We take the $x$,$y$, and $z$ coordinates of each mesh point instead of the cell as they do not have coordinates of their own. A point is a place of contact of cells with its neighbor. Thus we need to convert the cell data into point data. This is done by averaging the values of all cells attached at the point of contact. 

The meshes were huge, about 40 Megabytes each, and are densely packed, having around twenty-seven thousand nodes. To bring our computation time and cost further down, we use just the nodes on the external surface as the "wear" of a particular area is an external feature. 

\begin{figure}[h]
    \centering
    \includegraphics[width=12cm, height=3.5cm]{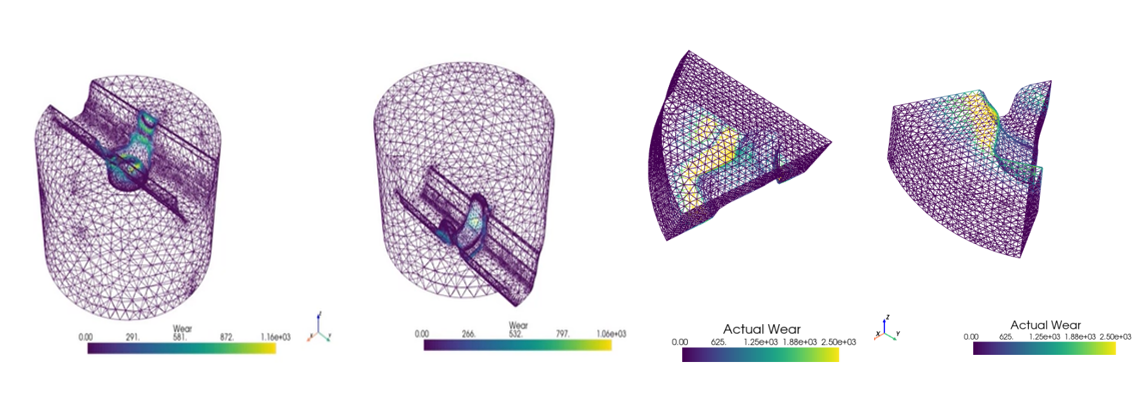}
    \caption{External surface of the mesh}
    \label{Fig 2}
\end{figure}

It can be seen in Fig~\ref{Fig 2} that many nodes in the mesh have zero "wear". The yellow color represents maximum wear and the purple zero wear. Having a sparse output vector with a skewed dataset could affect our deep learning model. Thus, taking only the external surface and also selecting the training initial conditions a bit far from one another helps better fit the model. After considering just the surface points, the upper deformable die now consists of around seven thousand points, and the lower deformable die is about nine thousand points. We have a total of 40 meshes in our dataset.

\begin{equation}
     G(V,E) \gets V, E list
\end{equation}

We initialize the node vector $V$ consisting of node features $ (nf_1, nf_2, nf_3, nf_4, nf_5) $, such as x,y, and z coordinates, with initial parameters as temperature, and friction coefficient for which wear is needed to be calculated.  The edges of the mesh are converted into an adjacency list  using cells $ [(n_1,n_3),(n_4, n_{10}), etc.]$ called $E list$. We use them to create the graph to be used as input to the model.

Similarly, we later tested for a new dataset of meshes, to check the network's credibility on new data. The new mesh is comparatively smaller, about 3 Megabytes each. There are about two thousand nodes in the external surface of the mesh. There are a total of sixty-four meshes in the new dataset. 
\subsection{Deep Learning Models}

The GNN model designs are used as a surrogate model to predict the final mesh with wear features. We have used PointNet and a GNN model with graph convolution layer \cite{shivadity2022graph} as a baseline model. 

There are two main network architectures used. We first have a model consisting of five edge convolution layers, followed by Rectified Linear (ReLU) activation function to add non-linearity to the layers. Since "wear" can only be positive or zero, thus using a ReLU function helps us as it only allows values greater than or equal to zero to pass to the next layer by deactivating the neuron with negative output, thus training the model faster.
The convolution is performed on the node features while also taking into account its neighboring node features. 

The graph and node feature vector are both given to the model. We first upscale the features to fifty and then to a hundred, followed by a fully connected layer. We then downscale these features back to fifty and, finally one. Thus at the end, we have a tensor of size equal to the number of nodes. Each node is associated with a value, which is then stored as a mesh feature "wear".

\begin{figure*}[h]
    \centering
    \includegraphics[width=12cm, height=2cm]{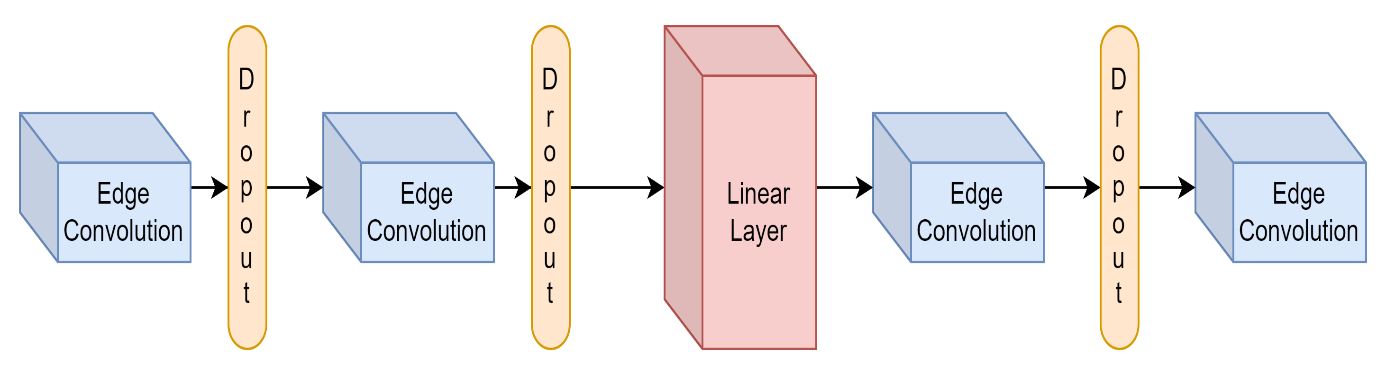}
    \caption{GNN model using edge convolution and liner layers}
    \label{Fig 3}
\end{figure*}

In the other model, we introduce a linear layer instead of the fully connected convolution layer, as shown in Fig~\ref{Fig 3}. As with only convolution layers, there is not much learning happening in the model. Also by adding a linear layer, we try to optimize a system of linear equations, similar to FEM.

On adding linear functions in between the convolutions with  different positional combinations, the best position was to replace it in the between instead of the fully connected convolution layer. This not only increased the accuracy slightly but also did not require a lot of additional training time, since the total number of layers is still the same. The linear layer has input and output dimensions equal to nodes in the graph by taking a transpose of the output vector of the convolution layer. We now have a linear equation for each node, with all its convoluted features. An increase in the number of  features in the liner layer did not lead to any more increase in accuracy.    

Dropout layers~\cite{srivastava2014dropout} are used in between the convolution and linear layers to regularise the model, by preventing over-fitting. We are trying to create a generalized model,  to make sure we do not over-fit the model when trained on a different set of data. By adding a dropout layer we randomly drop out or ignore some output node value, thus each layer is now different. It also helps to make the model more is robust by making the network adapt to correct mistakes from previous layers as each time there is a random dropout.

Different gradient descent algorithms are used to optimize the objective function consisting of model parameters by minimizing the error. The parameters are updated in the opposite direction of the gradient to reach a local or global minimum~\cite{ruder2016overview}. 
Adam optimization~\cite{kingma2014adam} gives better results than with just stochastic gradient~\cite{ruder2016overview} as it has a different learning rate associated with each parameter unlike in stochastic gradient. It is important to have a low learning rate and weight decay to not over-fit the model too early in the iteration, and to allow it to reach its correct minimum. For backpropagation, the Mean Absolute Error (MAE)~\cite{wang2018analysis} and the Mean Square Error (MSE)~\cite{wang2018analysis} are calculated over an iteration over a single node.

The other two models tested were of the same architecture as shown in Fig~\ref{Fig 3}, but we replace the edge convolution layer with the SAGE convolution layer. The model is built and trained using the deep graph library~\cite{wang2019deep} with PyTorch~\cite{paszke2019pytorch} backend.

\section{Results and Discussions}
\label{sec:results}

In this section, we discuss the criteria used to compare the models and analyze the results of our main models for both the upper and lower deformable die. In Table \ref{table:1}, the error percentage is represented to check the performance of the models.  The error percentage is calculated as

\begin{equation}
   Error \%  = {\frac{MAE}{Mean_{wear}}}  
\end{equation}

\begin{table}[h]
\caption{ Error percentage on the old and new dataset }
\label{table:1}
\begin{center}
\begin{tabular}{||c | c | c | c | c ||} 
 \hline
Model & \multicolumn{2}{|c|}{Dataset 1}& \multicolumn{2}{|c|}{Dataset 2
}\\ 

 \hline
 M & lower DD & upper DD & lower DD & upper DD  \\  
 \hline
 Mean & 90.82 & 48.34 & 305 & 265 \\
  \hline
 Maximum & 1100 & 857 & 4105 & 4162\\
\hline
 Graph Convolution & 39 \% & 65 \%  & 21.3 \% & 11.3 \% \\
 \hline
PointNet & 34 \% & 32.2 \% & 8.1 \% & 1.8 \% \\ 
 \hline
Edge Convolution(L) & 9.3 \%  & 13 \%  & 6.5 \% & 1.8 \% \\
 \hline
SAGE convolution(L) & 8.8 \%  & 13 \%  & 2.5 \% & 1.5 \% \\
 \hline
 
\end{tabular}
\end{center}
\end{table}

The error percentage may seem high, but it is because the mean is very low compared to higher points. This is due to the sparsity of the output vector nodes with zero wear. Table \ref{table:1} also shows that even though the error was low for the upper deformable die compared to the lower one, the error percentage lets us know that is due to the overall values and the mean, in general being low. Both the SAGE and edge convolution model with the linear layer have performed very well.

For the new dataset, although both the edge and SAGE models have the best performance, PointNet has also performed quite well compared to the graph convolution. This could mean that the graph convolution network is less susceptible to changes, that is the network parameters need to be optimized again for the new dataset, which is a time-consuming process. The PointNet trains the fastest followed by GNN, SAGE, and EDGE models, which have a similar training time. Calculating the error percentage helps to better understand the results in terms of the value for the company. 

\begin{figure}[h]
    \centering
    \includegraphics[width=12cm, height=5cm]{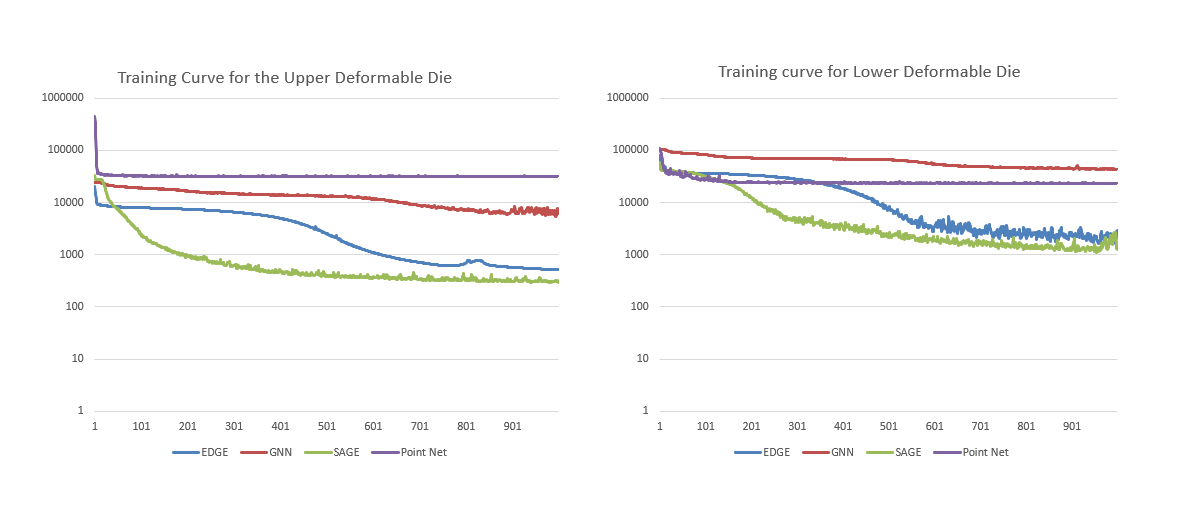}
    \caption{Learning curve for our main models}
    \label{Fig 4}
\end{figure}

The learning curve is plotted over the logarithm of average MSE error over all the training models. The learning rate for the upper deformable die is more smooth than that for the lower die shown in Fig~\ref{Fig 4}. The loss decreases quickly at the start except for the point cloud model and then after 700 epochs, the curve is still decreasing but at a very low pace. We could get a smoother curve with a smaller learning rate. The SAGE overall is smoother compared to the edge convolution model. We have run the model for about a thousand epochs to test for model stability. This also allows us to understand if there is a possibility of attaining a global minimum or local minimum. 

The model created can be used for any similar kind of mesh data structure, without having to change the code. The number of neurons in the linear layer depends on the input dimension of nodes in the graph, which is automatically adjusted. The optimization and the model initialization parameters are the same for all four types of meshes studied. The only drawback is that sometimes the learning curve may not be smooth and very rarely it would get stuck at local minima.

\subsection{Results for EDGE and SAGE model convolution with linear layer}

\begin{figure}[h]
    \centering
    \includegraphics[width=10cm, height=6cm]{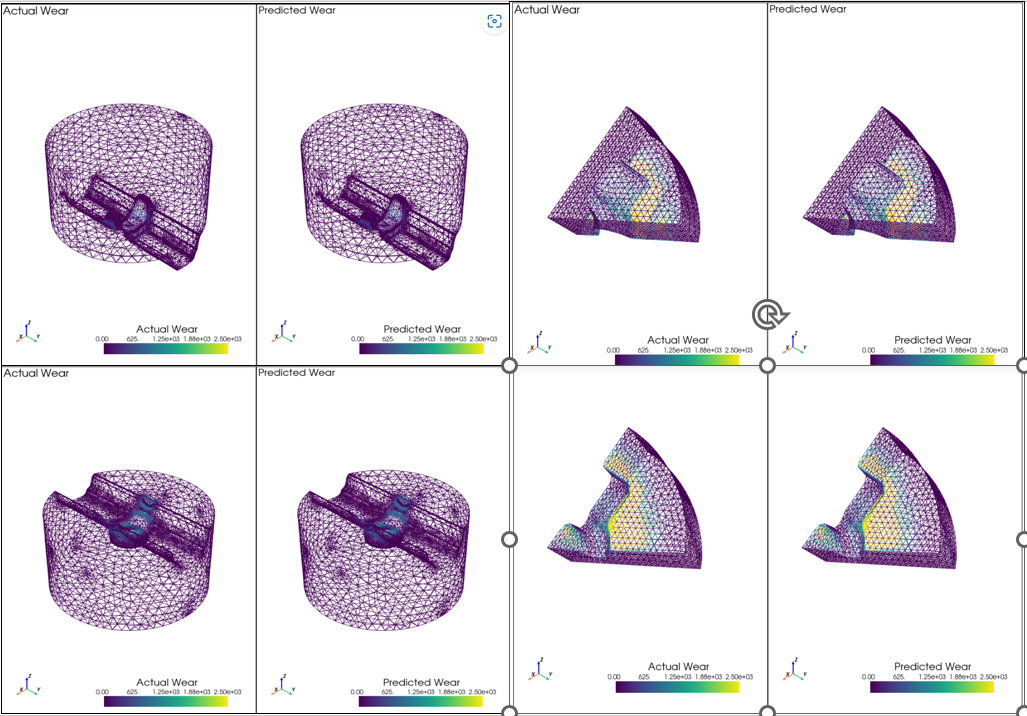}
    \caption{Comparison of actual and predicted meshes for both old and new dataset}
    \label{Fig 5}
\end{figure}

From the 3D figure representation of wear values at each node shown in Fig~\ref{Fig 5}, we see that the prediction is quite similar to the actual value. The pattern of the node values has been matched, with slight over-predictions at some points. We were mostly able to correctly detect the nodes with zero "wear" and the area of maximum wear. Though the values are not perfectly matched, the results are good and we are able to understand the "wear" distribution correctly over the die. Since the new die is a cylindrical sector instead of a cylinder, the mesh seen from different sides, is different, unlike the previous case when all the meshes were cylindrical. We can thus conclude the model is robust. 

\subsection{Time Cost Analysis for the hybrid process}

 Generating a thousand simulations with FEM would take around twenty days, can now be reduced to two to three days using the hybrid approach. We only need FEM simulations for around twenty to thirty initial conditions and the rest can be generated using the trained model. Thus, a cheaper time cost of about 85 \% could be achieved.  The trained model takes only 300 milliseconds to predict a new set of initial conditions. We could thus check for more initial conditions.

The time required by the neural network is proportional to the number of nodes in the mesh, as that would mean more features and parameters to be optimized. For training the upper die, the same network would take around three hours but for the lower die, it would be around five hours. It is similar for the completely new dataset as well. Since the new meshes are comparatively small, the network requires around one and a half hours to train. 

\section{Conclusion and future work}
\label{sec:Conclusion}

The main conclusion is that adding a linear layer to the model has increased the final accuracy. This might be due to trying to replicate a system of linear equations which is similar to the output of a FEM. For a neural network, accuracy increases with depth, but this would only be valid provided we have a large data source. In our case increasing layers increases the loss, as there are many new parameters to be calculated for the  new layers, but with fewer data which leads to more complexity. If initially we had a large number of unique features then maybe increasing the number of neurons might have led to better accuracy.

Hyperparameter tuning is very important as there have been large differences between the same model results with slight changes in the initializing model and optimization parameters. Adding a lot of convolution in the graph neural network has not increased accuracy to a great extent. Thus, maybe graph neural networks need not be too deep.

Future work was recently discussed with the company to run the model on more data-set with more initial parameters for the die. By increasing the sample in the training data-set, we could observe the changes in the error to check if the accuracy would increase, or if it over-fits the model. We also need to check for more completely new mesh shapes, to check for the generality of the model.

A check on the MSE loss after every 100 epochs can be made. If the learning curve is still following a decreasing trend, then continue the process. If it is almost constant with very small fluctuations around the mean, stop the process, as it means we have reached our optimization minima. Also, a different combination of graph convolutions and graph attention layers can be used to create a new model. More structural features could be extracted from the graph. It is important to make sure that these features are not too correlated with each other, as in that case, it would decrease the accuracy of the model.

\section*{Acknowledgment}

We thank Dr. Jose Alves, Scientific Developer at Transvalor S.A for providing the datasets for conducting this research.

\bibliographystyle{unsrt}
\bibliography{simple}

\end{document}